\documentclass[letterpaper, 10 pt, conference]{ieeeconf} 

\IEEEoverridecommandlockouts

\usepackage{epsfig} 
\usepackage{times} 
\usepackage[cmex10]{amsmath} 
\usepackage{amssymb}  

\usepackage{amsfonts}
\usepackage{hyperref}
\usepackage{cite}
\usepackage{graphicx}
\usepackage{algpseudocode}
\usepackage{algorithmicx}
\usepackage{color}
\usepackage{cases}

\algblock{ParFor}{EndParFor}
\algnewcommand\algorithmicparfor{\textbf{parallel for}}
\algnewcommand\algorithmicpardo{\textbf{do}}
\algnewcommand\algorithmicendparfor{\textbf{end\ parallel for}}
\algrenewtext{ParFor}[1]{\algorithmicparfor\ #1\ \algorithmicpardo}
\algrenewtext{EndParFor}{\algorithmicendparfor}

\newcommand{\rbb}{\mathbb{R}}

\newcommand{\p}[3][]{p\ifx\relax#1\relax\else_{#1}\fi\ifx\relax#2\relax\else^{#2}\fi\ifx\relax#3\relax\else\!\left( #3 \right)\fi}
\newcommand{\hp}[3][]{\widehat{p}\ifx\relax#1\relax\else_{#1}\fi\ifx\relax#2\relax\else^{#2}\fi\ifx\relax#3\relax\else\!\left( #3 \right)\fi}

\title{\LARGE \bf
Visual end-effector tracking using a 3D model-aided particle filter\\for humanoid robot platforms}

\author{Claudio Fantacci$^{1}$, Ugo Pattacini$^{1}$, Vadim Tikhanoff$^{1}$ and Lorenzo Natale$^{1}$%
\thanks{$^{1}$Claudio Fantacci, Ugo Pattacini, Vadim Tikhanoff and Lorenzo Natale are with Istituto Italiano di Tecnologia, iCub Facility, Humanoid Sensing and Perception, Via Morego 30, Genova, Italy {\tt\small claudio.fantacci@iit.it}, {\tt\small ugo.pattacini@iit.it}, {\tt\small vadim.tikhanoff@iit.it}, {\tt\small lorenzo.natale@iit.it}}%
}

\begin{document}
\maketitle
\thispagestyle{empty}
\pagestyle{empty}

\begin{abstract}
This paper addresses recursive markerless estimation of a robot's end-effector using visual observations from its cameras.
The problem is formulated into the Bayesian framework and addressed using Sequential Monte Carlo (SMC) filtering.
We use a 3D rendering engine and Computer Aided Design (CAD) schematics of the robot to virtually create images from the robot's camera viewpoints.
These images are then used to extract information and estimate the pose of the end-effector.
To this aim, we developed a particle filter for estimating the position and orientation of the robot's end-effector using the Histogram of Oriented Gradient (HOG) descriptors to capture robust characteristic features of shapes in both cameras and rendered images.
We implemented the algorithm on the iCub humanoid robot and employed it in a closed-loop reaching scenario.
We demonstrate that the tracking is robust to clutter, allows compensating for errors in the robot kinematics and servoing the arm in closed loop using vision.
\end{abstract}


\section{INTRODUCTION}
Humanoid robots are designed to work in an unstructured and possibly complex environment where humans live \cite{met2010,tsa2016} and they rely on a wide variety of sensors to perceive and interact with it.
The perception and interaction capabilities of a humanoid robot with such complexities depend on the available sensors.
However, the proprioception of the robot might be affected by a number of impairments, including measurement noises, sensor biases, mechanical interplay of the links, frictions and so forth.
This is particularly relevant for humanoid robots in which elastic elements (like tendon-driven actuators) introduce errors in the computation of the direct kinematics.
Precise estimation of objects using vision is also affected by errors introduced by imprecise knowledge of the cameras extrinsic parameters, especially in case in which cameras are moving to simulate the human oculomotor system.
As a consequence, tasks like reaching, grasping and, in general, manipulation with the end-effector can be problematic.
Vision plays an important role in this respect, as it mimics the human sense of vision and provides contactless measurements of the environment.
In this context, \textit{visual servoing} \cite{esp1992,hut1996,mal1999,kra2002,cha2006,cha2007} can be profitably employed to compensate for errors in the visual domain resorting to a closed loop control of the robot's end-effector by means of a visual feedback.
To control the robot's end-effector pose relative to a desired target, it is of central importance to have a precise knowledge of its pose over time.

Recursive Bayesian estimation is a well known tool for tracking an object/objects by fusing measurements from sensors \cite{far1985v1,bar1988,bar1995,bar2001}. In particular, \textit{particle filters} \cite{gor1993,dou2000,aru2002,ris2004} have proven to be effective in robotics \cite{thr2005} and in a wide variety of fields \cite{dou2001}.
Object tracking using visual feedback turns out to be challenging in that it generally requires the development of suitable filtering techniques that need to account for the presence of cluttered background and to be sufficiently fast to cope with the real-time specifications of the task.
To fulfill such requirements, Graphics Processing Unit (GPU) technologies, comprising their software development kit (e.g. CUDA \cite{cuda}, OpenCL \cite{opencl}), enable real-time implementation of complex recursive filtering pipelines for visual object tracking.

In this paper we propose a novel particle filter that estimates the 6D pose (considering position and orientation) of the robot's end-effector without the use of markers.
To yield our estimates, we resort to Computer Aided Design (CAD) schematics in order to render 3D mesh models of the end-effector as they would appear from the robot's viewpoints.
Finally, we extract Histogram of Oriented Gradient (HOG) \cite{dal2005} descriptors from the camera images that we compare against the rendered images.
We demonstrated that our approach is effective and robust in experimental visual servoing tests carried out in real settings using the iCub humanoid robot \cite{met2010}.

The rest of the paper is organized as follows.
Section \ref{sec:rel} provides an overview of visual hand tracking for both human and robotic hand.
Section \ref{sec:tracker} briefly introduces recursive Bayesian filtering and the particle filter formulation, then details our contributions to 6D pose estimation of a robot's end-effector using histogram of oriented gradient descriptors.
In Section \ref{sec:exp} we report on the experiments to validate our framework.
Finally, Section \ref{sec:con} provides concluding remarks and future work.


\section{RELATED WORK}
\label{sec:rel}
The problem of hand tracking has been extensively addressed in computer vision, especially for building intuitive interfaces.
Due to its similarity with the task we addressed in this paper we revise some of this work.

One of the first contribution on hand-shaped object tracking was in 1994 with DigitEyes \cite{reh1994}, which tackled the problem with a disjoint hand model-based approach (i.e. each part of the hand was considered an independent entity to be estimated), using a combination of line and point features with a Gauss-Newton algorithm \cite{boy2004}.
This method, however, does not perform well in complex scenario, such as cluttered background.

Later in 2000, another approach for hand tracking is presented under the name of Conditional Density Propagation (CONDENSATION) \cite{isa1998}.
CONDENSATION is based on a joint hand model-based approach and Sequential Monte Carlo (SMC) methods to fit a B-spline around a single hand in images.
This approach is capable of working in cluttered environment, although it does not provide 3D information about the pose as it can track only planar curves, and shows limitations when applied to rotating targets.
It is evident from \cite{isa1998} that SMC methods provide a reliable approach for estimating the state of an object in a cluttered scene.

In 2004, the Smart Particle Filter \cite{bra2004,bra2007} was proposed to track a hand with depth sensors and skin color detection, by using a Stochastic Meta-Descent (SMD) optimization method (based on gradient descent) and a particle filter to form what the authors refer to as ``smart particles''. The SMD optimization is used on each particle to explore the state space, then the new particles are combined together in a particle set by using importance sampling \cite{ris2004}. The method, although promising, makes use of depth sensors and skin color that can be unavailable on some humanoid robots; furthermore, the computational burden of using an optimization method and a particle filter may not fit well with the required time constraints.

More recently in 2006, the authors of \cite{ste2006} presented the Hierarchical Bayes Filter which is based on a tree-based filtering method that uses color and edge detectors.
This framework relies upon a tree of template hierarchy of the hand poses, which is manually built \cite{ste2004}, and uses skin color detectors which may lead to poor performance on a humanoid robotic platform.

In robotics the problem of hand localization has been often simplified using special markers. This approach requires to modify the robot, hardly scales to a cluttered environment and occlusions, but, more importantly, it does not allow to detect important parts of the hand like the tip of the hands or the torso. This is fundamental for grasping objects.
The first breakthrough dealing with humanoid robots and specifically tracking of an anthropomorphic hand are relatively recent, dating back to 2006 \cite{gra2011,fan2015,vic2015,vic2016}.

In \cite{gra2011} the authors proposed a novel framework based on the Virtual Visual Servoing (VVS) paradigm, first introduced in \cite{esp2002,com2006}.
3D CAD models are used within a rendering engine to virtually create the hand-effector as if it had been seen by a camera using information provided by the direct kinematics. The virtual end-effector is then compared against the real end-effector acquired by the camera images by means of the Chamfer distance transform \cite{bor1988} and its pose is refined via a classical visual servoing approach.

In \cite{vic2015,vic2016}, a computer graphics simulator is exploited to create the body schema of the robot, including an appearance model of the hand shape with texture. The output of the simulator is used to generate predictions about hand appearance in the robot camera images, based on the sensorimotor proprioceptive information, i.e. the motor encoders.
The predictions are then compared to the camera images using the Chamfer distance transform in a SMC algorithm to estimate the offset present in the robot's encoders.
It is worth pointing out that, in this framework, filtering is carried out on encoder biases which are then used to estimate the pose of the end-effector. This approach focuses on the estimation of robotic limb poses and may be problematic to be extended to other objects present in the field-of-view of the robot's cameras.

In this work, we proposes a recursive Bayesian filter using the VVS approach.
Our approach provides a novel particle filter architecture with the following contributions:  1$\left.\right)$ we render, for each particle, an image of the 3D mesh model of the end-effector as it would appear from the robot's viewpoints (likewise in \textit{augmented reality} contexts); 2$\left.\right)$ we then use this state representation to directly estimate the 6D pose (position and orientation) of the end-effector in the robot operative space using 2D image descriptors. In particular, we use HOG to compare the rendered images with the robot's camera images and, as a result, the particles which are more likely to represent the end-effector will have higher weight.
The main advantage of our approach is that, given proper 3D models, we can potentially track any other object in the robot operational space which is within the robot's camera field-of-view.
Visual servoing experiments with a real robotic platform show that our approach allows precise closed-loop control of the hand of the robot in cluttered environment and meets the time-constraints for a real-time implementation.


\section{6D END-EFFECTOR TRACKER}
\label{sec:tracker}
In this section, we briefly provide background on particle filtering and we detail our implementation and contributions.

\subsection{Particle filtering}
Filtering is the problem of recursively estimating over time $x_{k} \in \rbb^{n_{x}}$ of a dynamical system given the noisy measurement history $y_{1:k} \triangleq \left\{ y_{1}, \dots, y_{k} \right\}$, $y_{k} \in \rbb^{n_{y}}$.
In the Bayesian framework, the entity of interest is the \textit{posterior density} $\p[k]{}{x}$ that contains all the information about the state vector $x_{k}$ given all the measurements up to time $k$. Such a Probability Density Function (PDF) can be recursively propagated in time resorting to the well known Chapman-Kolmogorov equation and the Bayes' rule \cite{ho1964}
\begin{IEEEeqnarray}{rCl}
	\p[k|k-1]{}{x} & = & \int \varphi_{k|k-1}\!\left( x | \zeta \right) \p[k-1]{}{\zeta} d \zeta \, ,\label{eq:chapkol}\\
	\p[k]{}{x} & = & \dfrac{g_{k}\!\left( y_{k} | x \right) \, \p[k|k-1]{}{x}}{\displaystyle \int g_{k}\!\left( y_{k} | \zeta \right) \, \p[k|k-1]{}{\zeta} d \zeta} \, ,\label{eq:bayesrule}
\end{IEEEeqnarray}
given an \textit{initial density} $\p[0]{}{\cdot}$.
The PDF $\p[k|k-1]{}{\cdot}$ is referred to as the \textit{predicted density}, $\p[k]{}{\cdot}$ is the \textit{filtered density}, $\varphi_{k+1|k}\!\left( x | \zeta \right)$ is the \textit{Markov transition density} representing the conditional probability that the state at time $k+1$ will take value $x$ given that the state at time $k$ is equal to $\zeta$, and $g_{k}\!\left( y | x \right)$ is the \textit{measurement likelihood function} denoting the probability that the measurement at time $k$ will take value $y$ given the state $x$.

In many practical applications, such as navigation, tracking and localization, the transition and/or likelihood models are usually affected by nonlinearities and/or non-Gaussian noise distributions \cite{aru2002,gor1993}, thus precluding analytical solutions of eqs. (\ref{eq:chapkol}) and (\ref{eq:bayesrule}).
In these cases, one must invariably resort to some approximations.
Sequential Monte Carlo methods,  also known as \textit{particle filters}, can deal with arbitrary nonlinearities and distributions and supply a complete representation of the posterior state distributions.

The idea behind particle filtering is to approximate the posterior density at discrete-time $k-1$ by a set of random samples (particles) $\{w^{(i)}_{k-1},x^{(i)}_{k-1}\}_{1\leq i \leq N}$, where $x^{(i)}_{k-1}$ is the state of particle $i$, $w^{(i)}_{k-1}$ is its weight, i.e.
\begin{IEEEeqnarray}{c}
	\hp[k]{}{x} \approx \sum_{i=1}^{N}{w_{k}^{(i)} \, \delta(x - x_{k}^{(i)})} \, , \label{eq:pf} \\
	\sum_{i=1}^{N}w_{k}^{(i)} = 1 \, ,
\end{IEEEeqnarray}
where $\delta\!\left( \cdot \right)$ is the Dirac delta function.
This approximation of the posterior improves as $N\rightarrow \infty$ \cite{dou2000}.
In this way, the evaluation of the integrals of the Bayesian filtering equations (\ref{eq:chapkol}) and (\ref{eq:bayesrule}) is performed via the Monte Carlo numerical integration method, i.e., by transforming the integrals into discrete sums.

Given $\{w^{(i)}_{k-1},x^{(i)}_{k-1}\}_{1\leq i \leq N}$ and using the measurement $y_k$ at time $k$, the key is how to form the particle approximation of the posterior at $k$, i.e. $p_{k}(x)$ denoted as $\{w^{(i)}_k, x^{(i)}_k\}_{1\leq i \leq N}$.
In principle, the particle approximation (\ref{eq:pf}) can be computed by drawing a set of independent and identically distributed samples $x_{k}^{(i)}$, $i=1, \dots, N$, from the posterior $\p[k]{}{x}$. However, such a solution is not feasible because $\p[k]{}{x}$ is not known, thus the computation of the weights and particles at time $k$ is based on the concept of importance sampling \cite{rob2004}.
Let us introduce a \textit{proposal} or \textit{importance density} $\pi_{k}\!\left( x_{k} | x_{k-1}, \, y_{k} \right)$ to draw (preliminary) particles at time $k$:
\begin{equation}
	x^{(i)}_k  \sim  \pi_{k}\!\left( x | x^{(i)}_{k-1}, \, y_k \right) \, , \label{e:pf1}
\end{equation}
whose weights are computed as follows:
\begin{IEEEeqnarray}{rCl}
	\widetilde{w}^{(i)}_k & = & w^{(i)}_{k-1}
	\frac{g_k(y_k|x^{(i)}_k)\,\varphi_{k|k-1}(x^{(i)}_k|x^{(i)}_{k-1})}{\pi_k(x^{(i)}_{k}|x^{(i)}_{k-1},y_k)} \label{e:unnw}\\
	w^{(i)}_k & = &\frac{\widetilde{w}^{(i)}_k}{\sum_{j=1}^N \widetilde{w}^{(j)}_k} \label{e:pf3}
\end{IEEEeqnarray}
for $i=1,\dots,N$. This recursive procedure starts at time $k=0$ by sampling $N$ times from the initial PDF $p_{0}\!\left( \cdot \right)$.

The described particle method, also known as Sequential Importance Sampling (SIS), inevitably fails after many iterations, because all particle weights, except a few, become zero (a poor approximation of the posterior PDF due to particle degeneracy).
The collapse of the SIS scheme can be prevented by resampling the particles.
The resampling step chooses $N$ particles from $\{w^{(i)}_k, x^{(i)}_k\}_{1\leq i \leq N}$, where the selection of particles is based on their weights: the probability of particle $i$ being selected during resampling amounts to $w^{(i)}_k$.
After resampling, all particle weights are equal to $1/N$.
The simplest choice is to select $\pi_{k}\!\left( \cdot | \cdot \right)$ as the transitional density, i.e. $\pi_{k} \equiv \varphi_{k|k-1}$. In literature, this Particle Filter (PF) is also known as the \textit{bootstrap filter} \cite{gor1993}.

\subsection{Initialization and prediction step using direct kinematics}
To initialize the PF at time $k = 0$, each particle $x^{(i)}_{0}$ is set equal to the pose of the end-effector evaluated by means of the \textit{direct kinematics} map $f\!\left( q^{e}_{k} \right)$, where $q^{e}_{k} = \left\{ q^{e}_{k,\,1}, \dots, q^{e}_{k,\,n} \right\}$ are the joint angles, with $n$ the total number of joints of the kinematic chain connecting the base frame to the end-effector frame \cite{sic2010,sic2016}.

The direct kinematics map is also used to propagate particles over time: the joint angles are used to compute the motion of the end-effector $x_{k}$, i.e.
\begin{equation}
	x_{k} = f_{k-1}\!\left( x_{k-1}, q^{e}_{k-1}\right) + w_{k-1}
\end{equation}
where $f_{k-1}$ is provided by the Denavit-Hartenberg (DH) convention \cite{sic2010,sic2016} and $w_{k-1}$ is the process noise modeling uncertainties and disturbances in the object motion model.
As a result, the \textit{Markov transition density} is the PDF
\begin{equation}
	\varphi_{k|k-1}\!\left( x_{k} | x_{k-1},\, q^{e}_{k-1} \right) = \p[w]{}{x_{k} - f_{k-1}\!\left( x_{k-1}, q^{e}_{k-1} \right)} \, . \label{eq:markovpred}
\end{equation}

\subsection{Filtering step using CAD models and HOG descriptors}
We use a 3D rendering engine to virtually create images from the point of view of the robot camera depicting the mechanical model of the end-effector.
We then make use of HOG descriptors to extract features from them.
As a result, we can define a likelihood that compares the descriptors of the synthetic and real images.
For the sake of simplicity the following subsections consider a single camera viewpoint.

\subsubsection{Exploitation of CAD models}
the key idea is, at any time instant $k$, to virtually create an image $\widehat{I}_{k}$ displaying the mechanical structure of the manipulator as if it had been seen by the robot's camera.
This goal is achieved by means of a 3D rendering engine such, e.g., OpenGL \cite{shr2013}, Unity \cite{unity} or Unreal Engine \cite{unreal}, which are capable of carrying out multiple parallel rendering activities.
In order to avoid introducing dependencies and to have a code that is tailored for our applications, we decided to adopt the OpenGL solution.

\begin{figure}[thpb]
\centering
	\framebox{\parbox{0.97\linewidth}{\includegraphics[width=\linewidth]{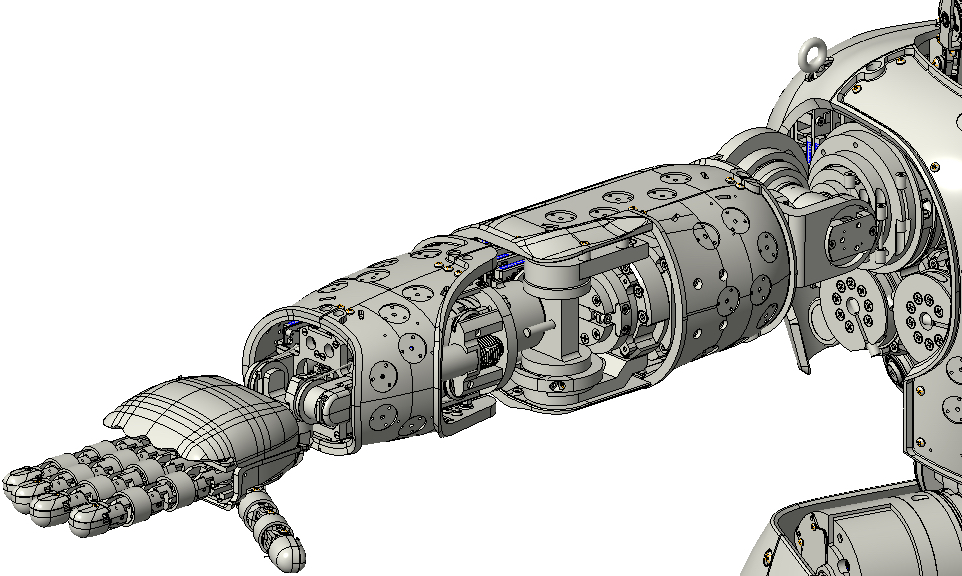}}}
        \caption{Mechanical structure of the right arm of the iCub humanoid robot platform}
        \label{fig:icubarm}
\end{figure}

\begin{figure}[thpb]
\centering
	\framebox{\parbox{0.8\linewidth}{\includegraphics[width=\linewidth]{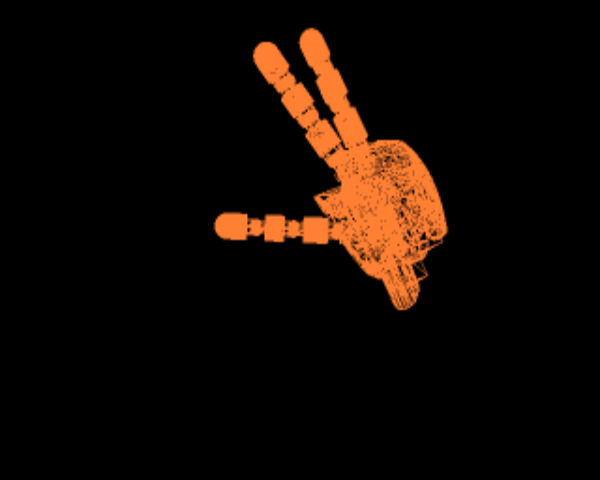}}}
        \caption{Rendered image of the right end-effector (hand) of the iCub humanoid robot. In the context of this work, we decided to disable the ring and little fingers from being rendered. Motivations are detailed in Section \ref{sec:exp}.}
        \label{fig:render}
\end{figure}

The mechanical structure of the manipulator chain of a humanoid robot consists of an arm and a hand, e.g., see Fig. \ref{fig:icubarm}.
To render the mechanical structure of the manipulator with respect to the reference frame of the robot and project it to the camera plane given the pose of the end-effector $x_{k}^{(i)}$, we need the joint angles $q^{e}_{k}$, the joint angles $q^{c}_{k}$ of the camera kinematic chain and the \textit{camera calibration matrix} $K$ accounting for the intrinsic camera parameters \cite{har2003}.
Formally, at any time instant $k$, the \textit{rendered} or \textit{predicted image} is defined as
\begin{equation}
	\widehat{I}_{k}^{(i)} \triangleq r\!\left( x_{k}^{(i)}, q^{e}_{k}, q^{c}_{k}, K \right) \, , \label{eq:rendering}
\end{equation}
where internally a roto-translation from a common reference frame to the camera image plane is performed by means of the projection matrix
\begin{equation}
	\Pi = K \, H\!\left( q^{c}_{k} \right) \, , \label{eq:proj}
\end{equation}
being $H\!\left( q^{c}_{k} \right)$ the homogeneous transformation from the common reference frame to the camera frame.
A pictorial representation of (\ref{eq:rendering}) is shown in Fig. \ref{fig:render}.

\subsubsection{Likelihood evaluation using HOG descriptors}
a \textit{measurement likelihood function} $g_{k}\!\left( \cdot | \cdot \right)$ needs to be defined to update the weights of the particles, using measurement $y_{k}$ extracted from the camera image $I_{k}$ and the information $\widehat{y}^{(i)}_{k}$ extracted from the rendered image $\widehat{I}^{(i)}_{k}$ given $x_{k}^{(i)}$.
The key idea is to extract image features characterizing the shape of the objects, so that the image $\widehat{I}^{(i)}_{k}$, associated to particle $i$, that is more similar to the image $I_{k}$ will be rewarded with a higher weight $\widetilde{w}_{k}^{(i)}$.

HOG descriptors capture robust characteristics of shape (gradient structure, edges) in a local representation with an easily controllable degree of invariance to local geometric and photometric transformations. The associated descriptors are computed on a dense grid of uniformly spaced cells, and provide robust information about shapes in the presence of a cluttered environment.
The HOG $h\!\left( \cdot \right)$ is compared to assess whether a rendered image $\widehat{I}^{(i)}_{k}$ is more likely to represent the real manipulator configuration captured by the camera image $I_{k}$.

Using the renderer $r\!\left( \cdot \right)$ and the HOG descriptors $h\!\left( \cdot \right)$ on both the camera and the predicted image allows defining the following likelihood function:
\begin{IEEEeqnarray}{rCl}
	g_{k}\!\left( y_{k} | x^{(i)}_{k} \right) & \triangleq & e^{-\dfrac{1}{\sigma}\left| y_{k} - \widehat{y}^{(i)}_{k} \right|} \nonumber \\
							     & = & e^{-\dfrac{1}{\sigma}\left| h(I_{k}) - h(\widehat{I}^{(i)}_{k}) \right|} \, ,\label{eq:hoglik}
\end{IEEEeqnarray}
where $\sigma$ is a free tuning parameter.

The pseudo code of the overall PF algorithm is reported in Table \ref{a:pf} for a single processing cycle at time instant $k$. 
\begin{table}[thpb]
	\caption{Particle Filter (PF)}
	\vspace{-2em}
	\label{a:pf}
	\hrulefill\hrule
	\begin{algorithmic}[0]
		\Function{Particle Filter}{$\{x^{(i)}_{k-1}, w^{(i)}_{k-1}\}_{1\leq i \leq N}$, $y_k$}
		\vspace{0.5em}
		\ParFor{$i=1,\dots,N$}\vspace{0.25em}
			\State \textsc{Draw a sample} $x^{(i)}_k \sim \varphi_{k|k-1}(x_k|x^{(i)}_{k-1}, \, q^{e}_{k-1})$ \vspace{1em} \Comment{See (\ref{eq:markovpred})}				\State \textsc{Render image} $\widehat{I}_{k}^{(i)} = r\!\left( x_{k}^{(i)}, q^{e}_{k}, q^{c}_{k}, K \right)$\vspace{1em}
			\State \textsc{Compute HOG} $\widehat{y}_{k}^{(i)} = h\!\left( \widehat{I}_{k}^{(i)} \right)$\vspace{-0.25em}
			\State \textsc{Update weight} $\widetilde{w}^{(i)}_k = w_{k-1}^{(i)} \, e^{-\dfrac{1}{\sigma}\left| y_{k} - \widehat{y}_{k}^{(i)} \right|}$\vspace{0.25em}
		\EndParFor\vspace{0.5em}
		\State $w_k^{(i)} = \widetilde{w}_k^{(i)}/\sum_{j=1}^N\widetilde{w}_k^{(j)}$, \textsc{for}  $i=1,\dots,N$
		\State \textsc{Compute weight degeneracy:} $\widehat{N}_{\mbox{\tiny eff}} = \left[\sum_{i=1}^N \left(w_k^{(i)}\right)^2 \right]^{-1}$
		\If{$\widehat{N}_{\mbox{\tiny eff}} < N_{\mbox{\tiny thr}}$}\vspace{0.25em}
			\State \textsc{Resample} $\{x^{(i)}_{k}, w^{(i)}_{k}\}_{1\leq i \leq N}$\vspace{0.25em} \Comment{See \cite[Table 3.2]{ris2004}}
		\EndIf
	\EndFunction\vspace{0.5em}
	\end{algorithmic}
	\hrule\hrulefill
\end{table}


\section{EXPERIMENTAL RESULTS}
\label{sec:exp}
To evaluate the effectiveness and robustness of the proposed approach for estimating the pose of the end-effector, a C++ implementation of the PF algorithm has been tested on the iCub platform \cite{met2010}.
We ran our experiments on a laptop with an Intel i7 3.40 GHz processor and an NVIDIA GeForce GT 750M with 2048 MB VRAM.
To implement (\ref{eq:hoglik}), we use the CUDA HOG implementation available in OpenCV \cite{opencv} as it provides a significant speed boost compared to the standard CPU implementation.

\subsection{Experimental setup}
The final goal of the experiments is twofold: 1$\left.\right)$ to assess the filtering capabilities of our approach and 2$\left.\right)$ to use it in visual servoing-aided reaching tasks.

\begin{figure}[thpb]
\centering
	\framebox{\parbox{0.8\linewidth}{\includegraphics[width=\linewidth]{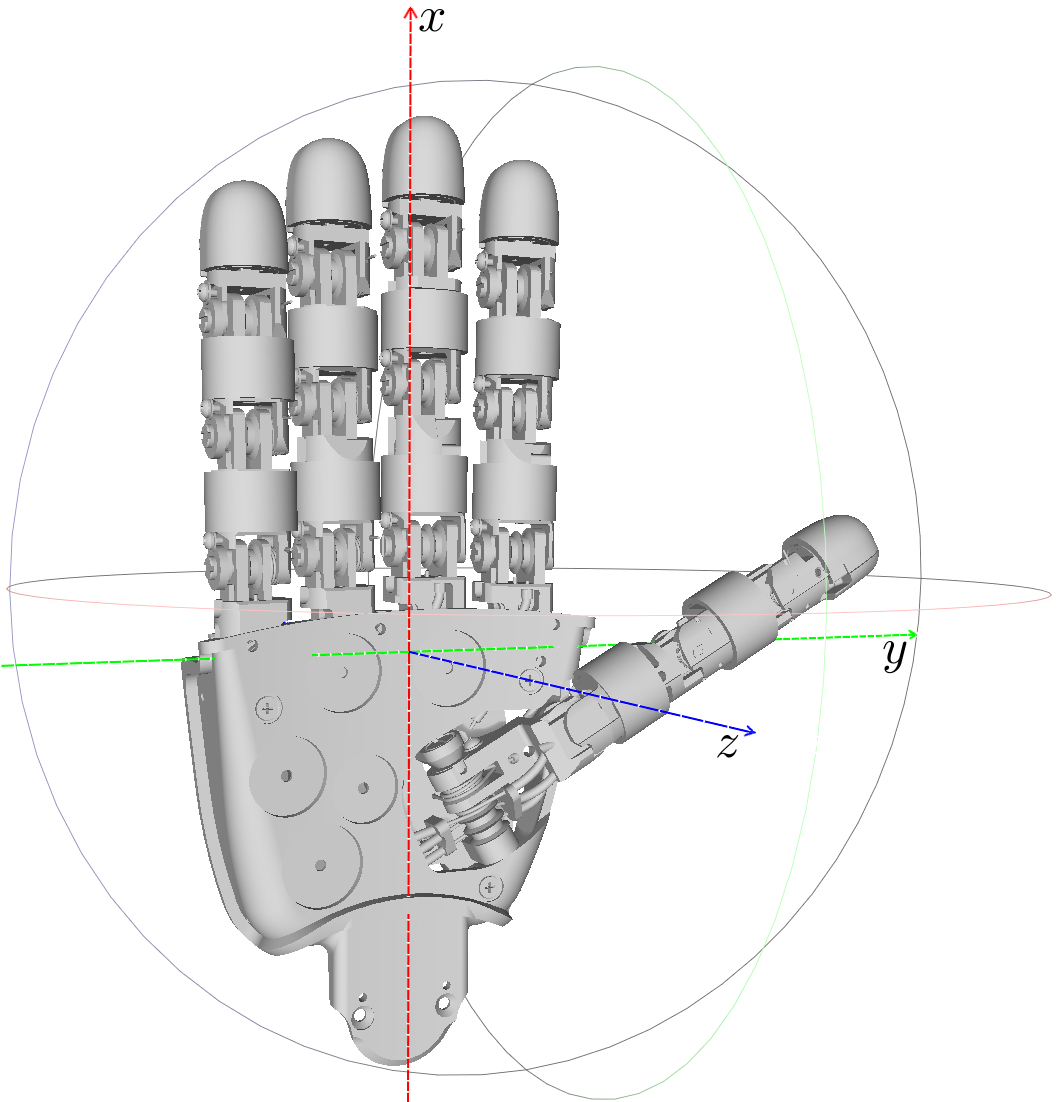}}}
        \caption{The iCub right hand.}
        \label{fig:hand}
\end{figure}

The end-effectors of the iCub are defined as a point on the surface of the left and right palm, below the middle fingers (a pictorial representation is provided in Fig. \ref{fig:hand}).
However, we can freely move the end-effectors to ease the execution of the task.
In the context of our experiments, we consider the right hand end-effector and we move it from the palm to the right index fingertip.

The state of the pose of the end-effector is denoted by $x = \left[ p_{x}, \, p_{y}, \, p_{z}, \, o_{x, \vartheta}, \, o_{y, \vartheta}, \, o_{z, \vartheta}, \, \right]^{\top}$, where:
$\left( p_{x}, \, p_{y}, \, p_{z}, \right)$ is the cartesian position of the end-effector with respect to the root reference frame which is positioned at the level of the waist in the center of the robot \cite{met2010};
$\left( o_{x, \vartheta}, \, o_{y, \vartheta}, \, o_{z, \vartheta} \right)$ is the orientation of the end-effector expressed with a compact axis-angle notation having the unit vector axis $o = \left[ o_{x}, \, o_{y}, \, o_{z} \right]^{\top}$ multiplied by the rotation angle $\vartheta$, i.e. $o_{x, \vartheta} \triangleq  \vartheta \, o_{x}$, $o_{y, \vartheta} \triangleq  \vartheta \, o_{y}$, $o_{z, \vartheta} \triangleq  \vartheta \, o_{z}$.
The motion of the particles is modeled according to eq. (\ref{eq:markovpred}), with $w_{k-1}$ defined as follows: the position disturbances are modeled as a white Gaussian noise with standard deviation $\sigma_{p} = 0.005 \, [m]$; the orientation disturbances are modeled as a Gaussian noise on a spherical cap \cite{har1998}, with standard deviations $\sigma_{\theta} = 3 \, [deg]$ and $\sigma_{\alpha} = 1.5 \, [deg]$ for, respectively, the rotation angle and the cap aperture. We use $N = 100$ particles to balance the performance and the computational burden of the PF, and $N_{thr} = 10$ for the resampling threshold.
All the PF parameters are summarised in Table \ref{tab:param}.
\begin{table}[thpb]
	\renewcommand{\arraystretch}{1.3}
	\caption{Parameter set for the PF}
	\label{tab:param}
	\centering
	\begin{tabular}{cc}
		\hline\hline
		$\sigma_{p}$ & $0.005 \, [m]$\\
		$\sigma_{\theta}$ & $3 \, [deg]$\\
		$\sigma_{\alpha}$ & $1.5 \, [deg]$\\
		$N$ & 100 \\
		$N_{thr}$ & $10$ \\
		\hline\hline
	\end{tabular}
\end{table}

During each rendering call, we use the CAD models of the right palm, thumb, index and middle fingers.
The iCub ring and little fingers are both coupled and underactuated by a single motor.
Due to the lack of a proper modeling of their kinematics, we could only get imprecise information about their pose.
As a consequence, we decided to disable these two fingers from being rendered.

\begin{figure}[thpb]
\centering
	\framebox{\parbox{0.97\linewidth}{\includegraphics[width=\linewidth]{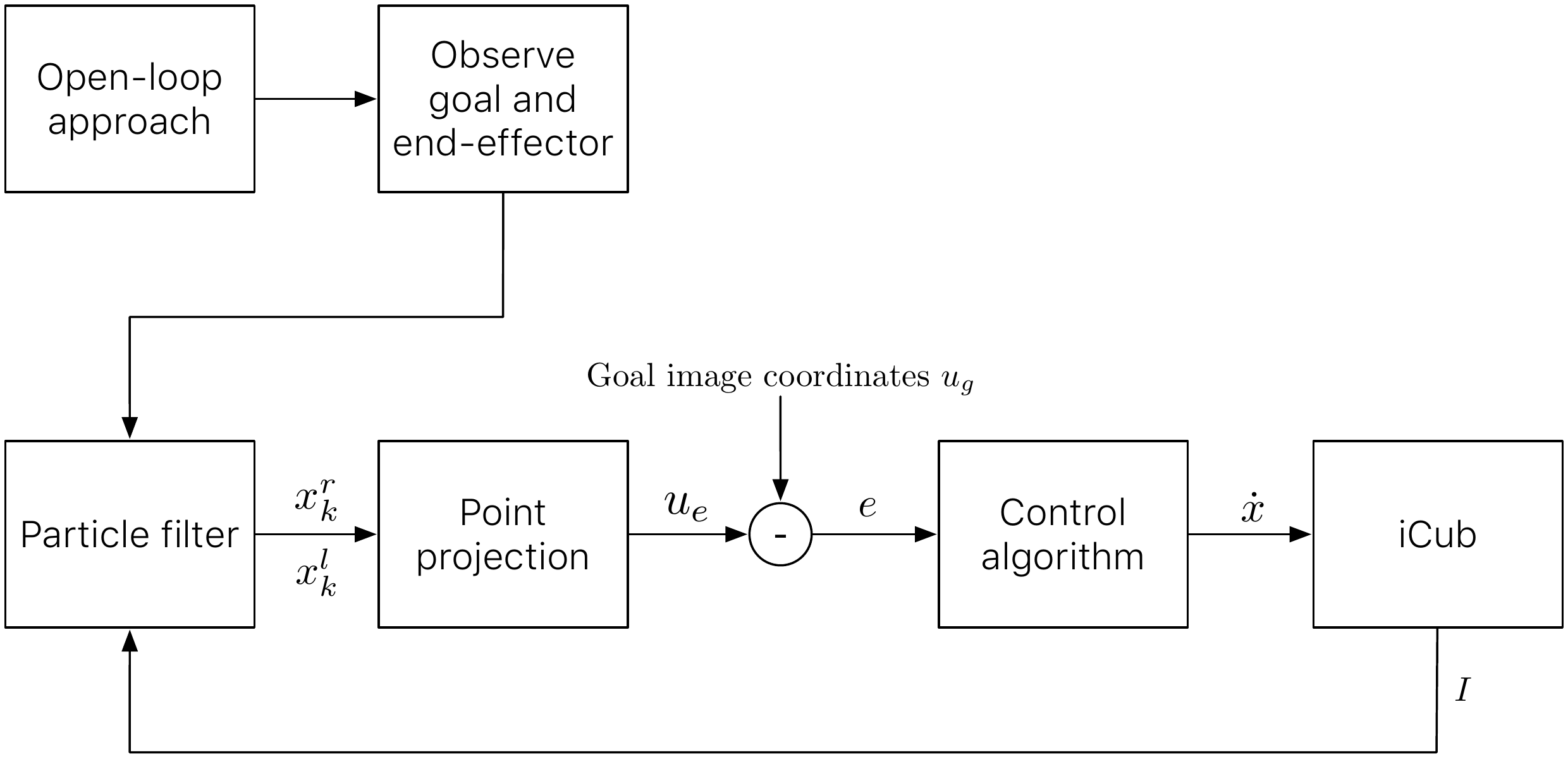}}}
        \caption{Pipeline of the two tasks.}
        \label{fig:pipeline}
\end{figure}

We planned two different reaching tasks on the iCub with two different goals: with the former task we aimed to demonstrate that our PF provides reliable estimates of the end-effector for visual servoing, achieving sub-pixel precision; the latter task was to show that we can perform precise reaching in a cluttered scenario.
The complete pipeline of our visual servoing loop is shown in Fig. \ref{fig:pipeline}.
\\
\textbf{Open-loop approach:} before entering the visual servoing control loop, we use the iCub stereo vision to get a rough 3D localization of our target. In particular, we employ a Structure From Motion algorithm \cite{fan2015} to get such 3D point and then we move the right hand using an open loop control.\\
\textbf{Observe goal and end effector:} in order to carry out visual tracking and visual servoing the robot has to observe both the target object and its end-effector. We then use the gaze controller of the iCub \cite{ron2016} to focus the attention of the robot to these two targets. This configuration is also known in literature to as \textit{endpoint closed-loop system} \cite{hut1996}.\\
\textbf{Particle Filter:} the PF described in section \ref{sec:tracker} is initialized and run for each camera. The output are the poses $x_{k}^{l}$ and $x_{k}^{r}$ of the right hand fingertip, respectively, for the left and right camera. Estimates are evaluated using a \textit{weighted mean} on the particle set, referred to as the Expected a Posterior (EAP) estimates \cite{sil1986}.\\
\textbf{Point projection:} estimates carried out by the PF are projected, respectively for each camera, onto the camera plane. This amounts to multiplying the position of the right hand fingertip $x_{k}^{l}$ and $x_{k}^{r}$ (in homogeneous form) by the left and right camera version of eq. (\ref{eq:proj}), which is equivalent to evaluate 
\begin{numcases}{}
	\lambda_{l} u_{l} = f_{1}\!\left( x, \, y, \, z\right) \label{eq:ul}\\
	\lambda_{l} v_{l} = f_{2}\!\left( x, \, y, \, z\right) \label{eq:vl}\\
	\lambda_{l} = f_{3}\!\left( x, \, y, \, z \right) \label{eq:ll}
\end{numcases}
for the left camera and
\begin{numcases}{}
	\lambda_{r} u_{r} = f_{4}\!\left( x, \, y, \, z\right) \label{eq:ur}\\
	\lambda_{r} v_{r} = f_{5}\!\left( x, \, y, \, z\right) \label{eq:vr} \\
	\lambda_{r} = f_{6}\!\left( x, \, y, \, z \right) \label{eq:lr}
\end{numcases}
for the right camera. Note that functions $f_{j}\!\left( \cdot \right)$, $j = 1, \dots, 6$, are linear in $x$, $y$ and $z$.
As a result, from eqs. (\ref{eq:ul})-(\ref{eq:ll}) and (\ref{eq:ur})-(\ref{eq:lr}), we collect the two coordinate pairs on the left and right image plane $\left( u_{l}, \, v_{l}\right)$ and $\left( u_{r}, \, v_{r}\right)$.
Finally, we define the goal image coordinate controlled by the visual servoing algorithm for the reaching task as $u_{e} \triangleq \left[ u_{l}, \, u_{r}, \, v_{l} \right]^{\top}$.\\
\textbf{Goal image coordinate:} we manually provide the goal image coordinate $u_{g} \triangleq \left[ u_{l}^{g}, \, u_{r}^{g}, \, v_{l}^{g} \right]^{\top}$, used to evaluate the reaching error $e \triangleq u_{g} - u_{e}$.\\
\textbf{Control algorithm:} the control algorithm takes the error $e$ as input and computes the cartesian velocities $\dot{x}$ to be performed by the end-effector to achieve $e \rightarrow \left[ 0, \, 0,\, 0 \right]^{\top}$.
To relate changes in point coordinates to changes in position of the robot, the image Jacobian $J$ is calculated from eqs. (\ref{eq:ul})-(\ref{eq:lr}).
We note that the Jacobian turns out to be a $3 \times 3$ matrix relating $\left[ x, \, y, \, z \right]^{\top}$ to $\left[ u_{l}, \, u_{r}, \, v_{l} \right]^{\top}$. 
The missing columns corresponding to the orientation of the end-effector are not considered because we make the simplifying assumption that the final desired orientation is the one provided by the open loop approaching phase.
The cartesian velocities are finally evaluated inverting the Jacobian matrix $J$ \cite{hut1996,fan2015} with
\begin{equation}
	\dot{x} = K_{\dot{x}} J^{-1} \dot{e} \, ,
\end{equation}
with $K_{\dot{x}} > 0$ a proportional gain, and are used by a \textit{cartesian controller} \cite{pat2010} implemented on the iCub platform.
The main advantage of using the cartesian controller is that it automatically deals with singularities and joint limits, and can find solutions in virtually any working conditions.\\
\textbf{iCub:} while motions are performed, new camera images are provided to the particle filter to re-iterate the pipeline.

\subsection{Task 1}
The goal of the first task is to assess whether we can perform visual servoing by using the estimates provided by our PF and achieve precise reaching of the goal.
The iCub is required to reach the fixed point $u_{g} = \left[ 125,\, 89,\, 135 \right]^{\top}$, from $10$ different starting points and with two different velocities, $0.005 \, [m/s]$ and $0.02 \, [m/s]$.
No objects are present in the path between the starting to the goal point.
The termination condition is achieved when the $\ell^{2}$-norm of $e$ falls below $1$ pixel.

\begin{figure}[thpb]
\centering
	\framebox{\parbox{0.97\linewidth}{\includegraphics[width=\linewidth]{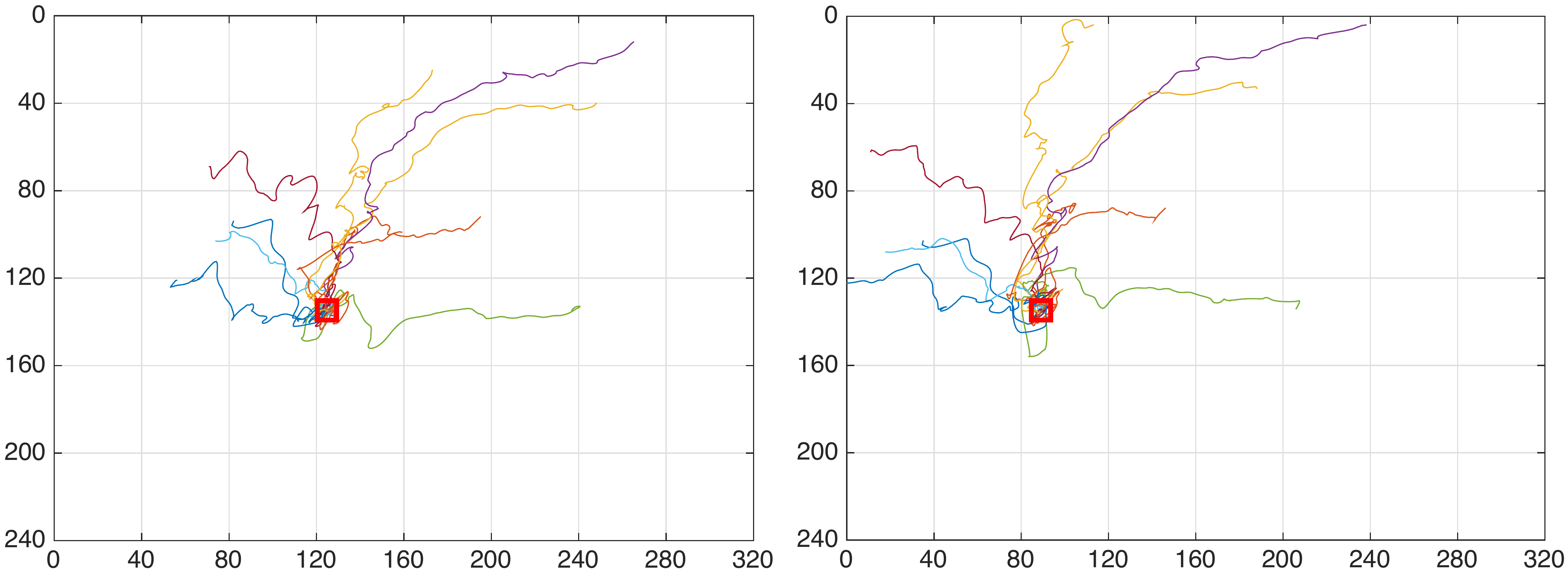}}}
        \caption{Trajectories of the end-effector, with $0.005 \, [m/s]$ velocity, seen from the left and the right cameras.}
        \label{fig:slow}\vspace{0.5em}
        \framebox{\parbox{0.97\linewidth}{\includegraphics[width=\linewidth]{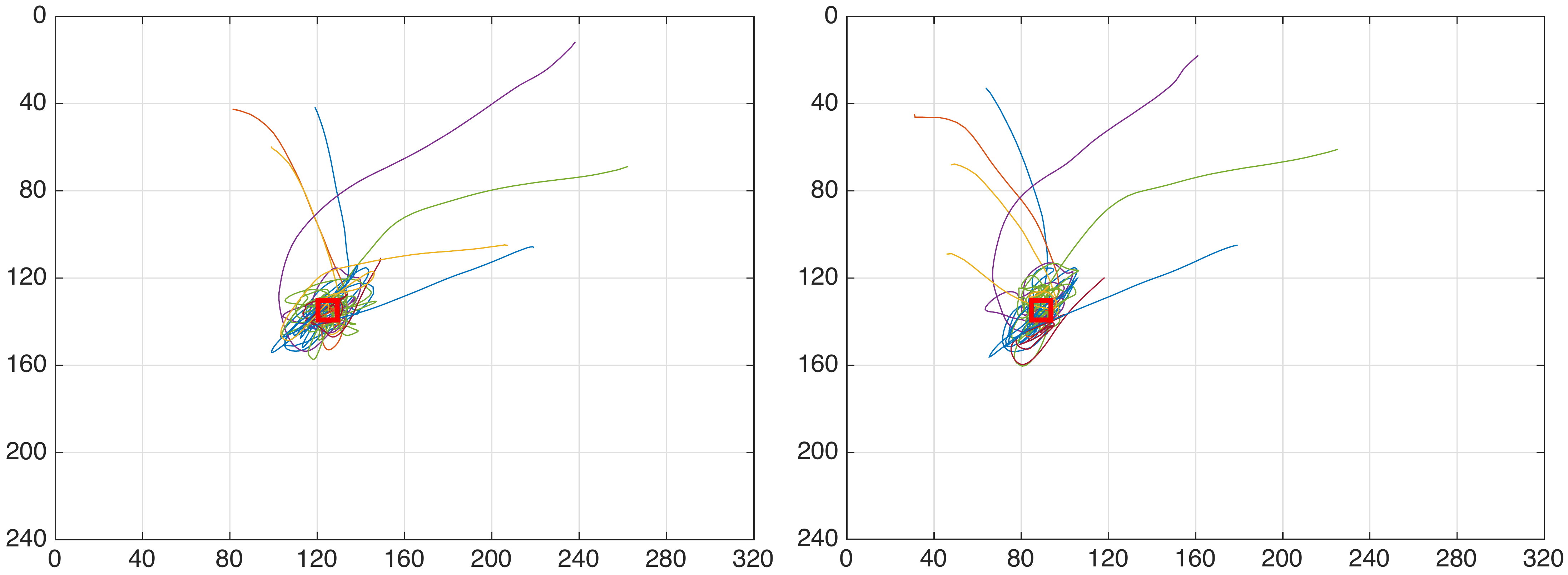}}}
        \caption{Trajectories of the end-effector, with $0.02 \, [m/s]$ velocity, seen from the left and the right cameras.}
        \label{fig:fast}\vspace{0.5em}
\end{figure}

Table \ref{tab:firsttask} summarises the results of the first task.
The robot achieved sub-pixel precision for all $20$ trials with a mean square error of $0.88 \pm 0.402 \, [pixel]$ and $0.977 \pm 0.104 \, [pixel]$ respectively for the slower and faster velocity.
\begin{table}[thpb]
	\renewcommand{\arraystretch}{1.3}
	\caption{Summary table of the the first task}
	\label{tab:firsttask}
	\centering
	\begin{tabular}{|c|c|c|}
		\hline
		\textbf{Velocity} & \textbf{Success/Trials} & \textbf{Mean square error}\\
		\hline\hline
		$0.005 \, [m/s]$ & $10/10$ & $0.88 \pm 0.402 \, [pixel]$ \\
		$0.02 \, [m/s]$ & $10/10$ & $0.977 \pm 0.104 \, [pixel]$ \\
		\hline
	\end{tabular}
\end{table}
A pictorial view of the trajectories are shown in Figs. \ref{fig:slow} and \ref{fig:fast}.
It is worth noticing that faster velocities produce very smooth trajectories, but longer time lapses are required to reach the goal.
On the other hand, slower velocities produce trajectories that are directed to the goal, but are more spiky.
While the spiky trajectories are due to the slow motion of the end-effector, the slow convergence to the goal of the faster trajectories can be mitigated by employing a higher number of particles and by refining the control algorithm.

\subsection{Task 2}
\begin{figure}[thpb]
\centering
	\framebox{\parbox{0.97\linewidth}{\includegraphics[width=\linewidth]{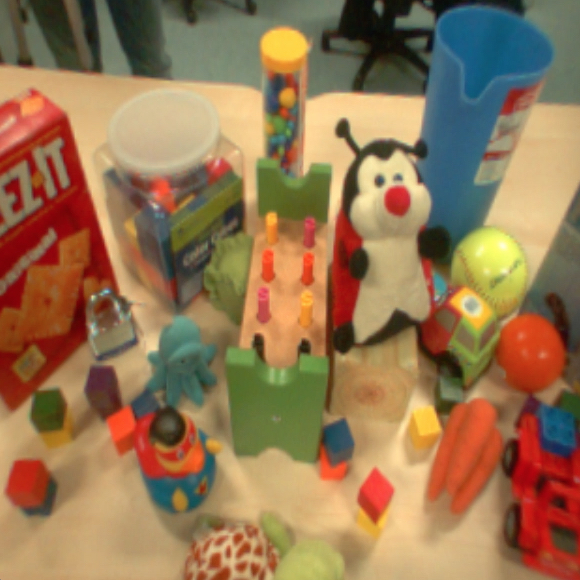}}}
        \caption{Image of the table from the iCub left camera viewpoint.}
        \label{fig:table}
\end{figure}

\begin{figure}[thpb]
\centering
	\framebox{\parbox{0.97\linewidth}{\includegraphics[width=\linewidth]{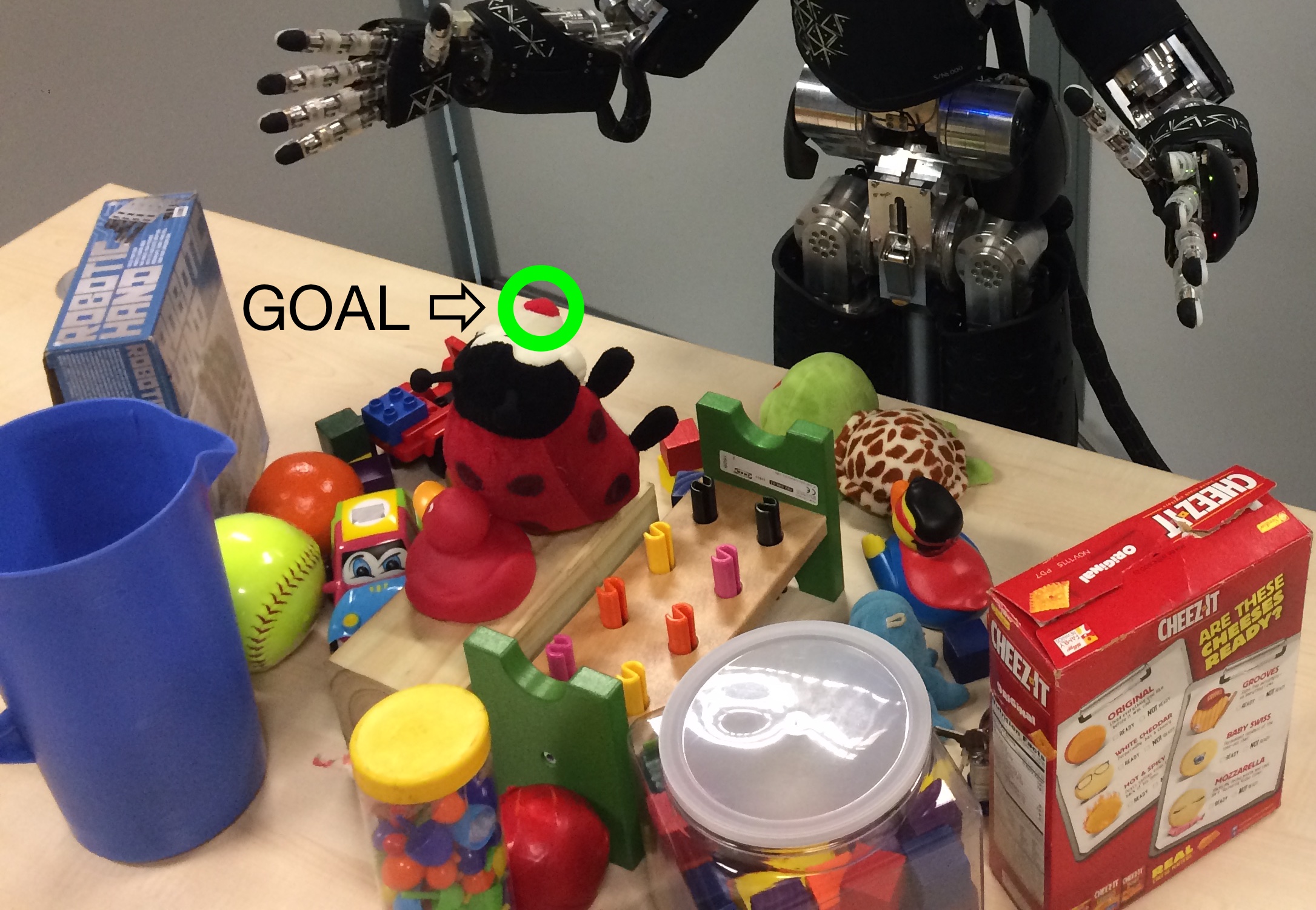}}}
        \caption{Pictorial view of the experiment setup. The ladybug red nose is highlighted by the green circle.}
        \label{fig:setup}
\end{figure}

For the second task, the iCub is presented with a table full of everyday objects and toys, as depicted in Fig. \ref{fig:table}.
In this settings, it is possible to test the robustness and effectiveness of the our PF.
The goal is to reach for the red nose of a ladybug plush, as highlighted in Fig. \ref{fig:setup}, for $10$ trials.

\begin{figure}[thpb]
\centering
	\framebox{\parbox{0.97\linewidth}{\includegraphics[width=\linewidth]{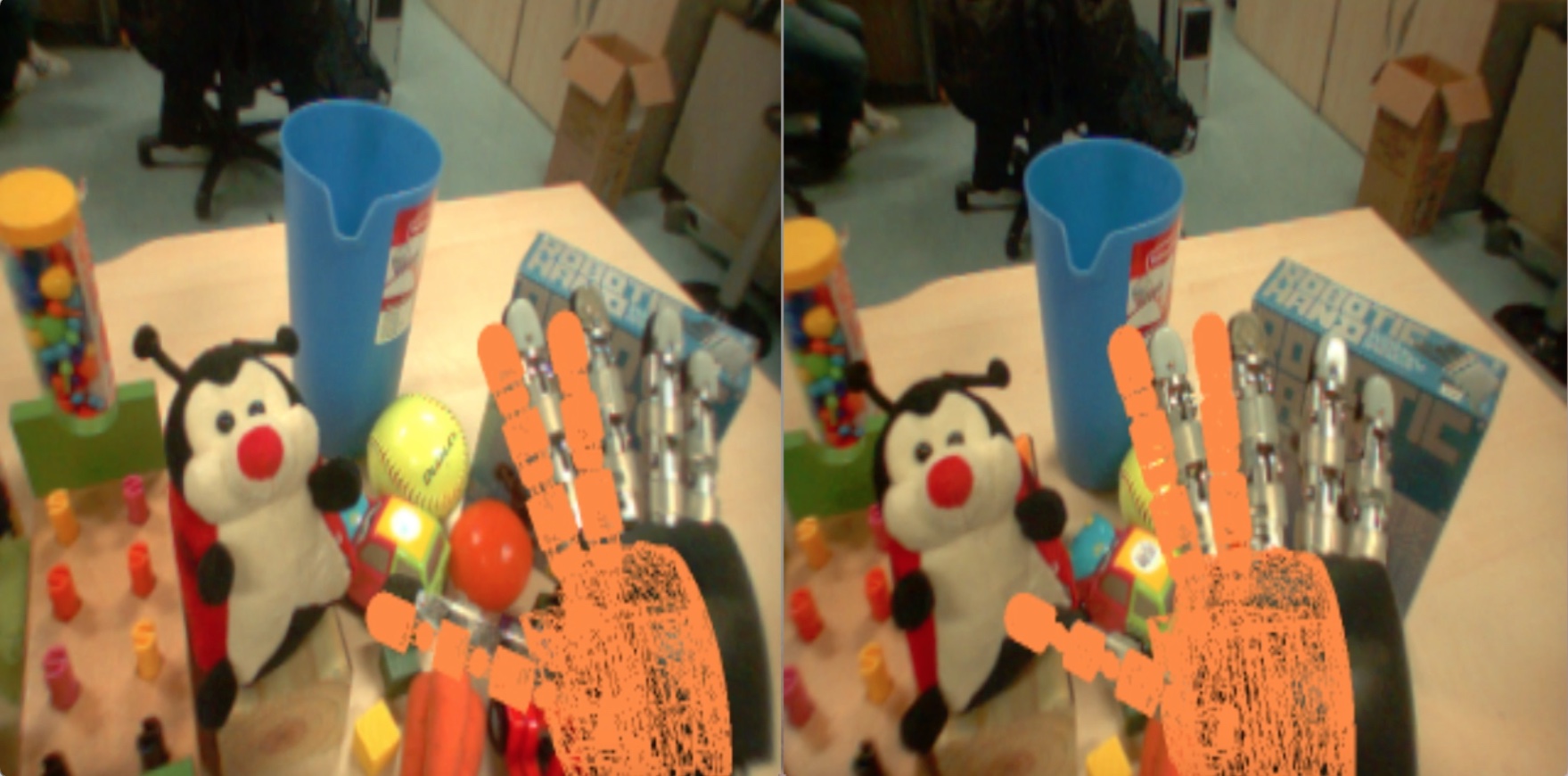}}}
        \caption{Error of the end-effector pose using direct kinematics.}
        \label{fig:reach1}\vspace{0.5em}
        \framebox{\parbox{0.97\linewidth}{\includegraphics[width=\linewidth]{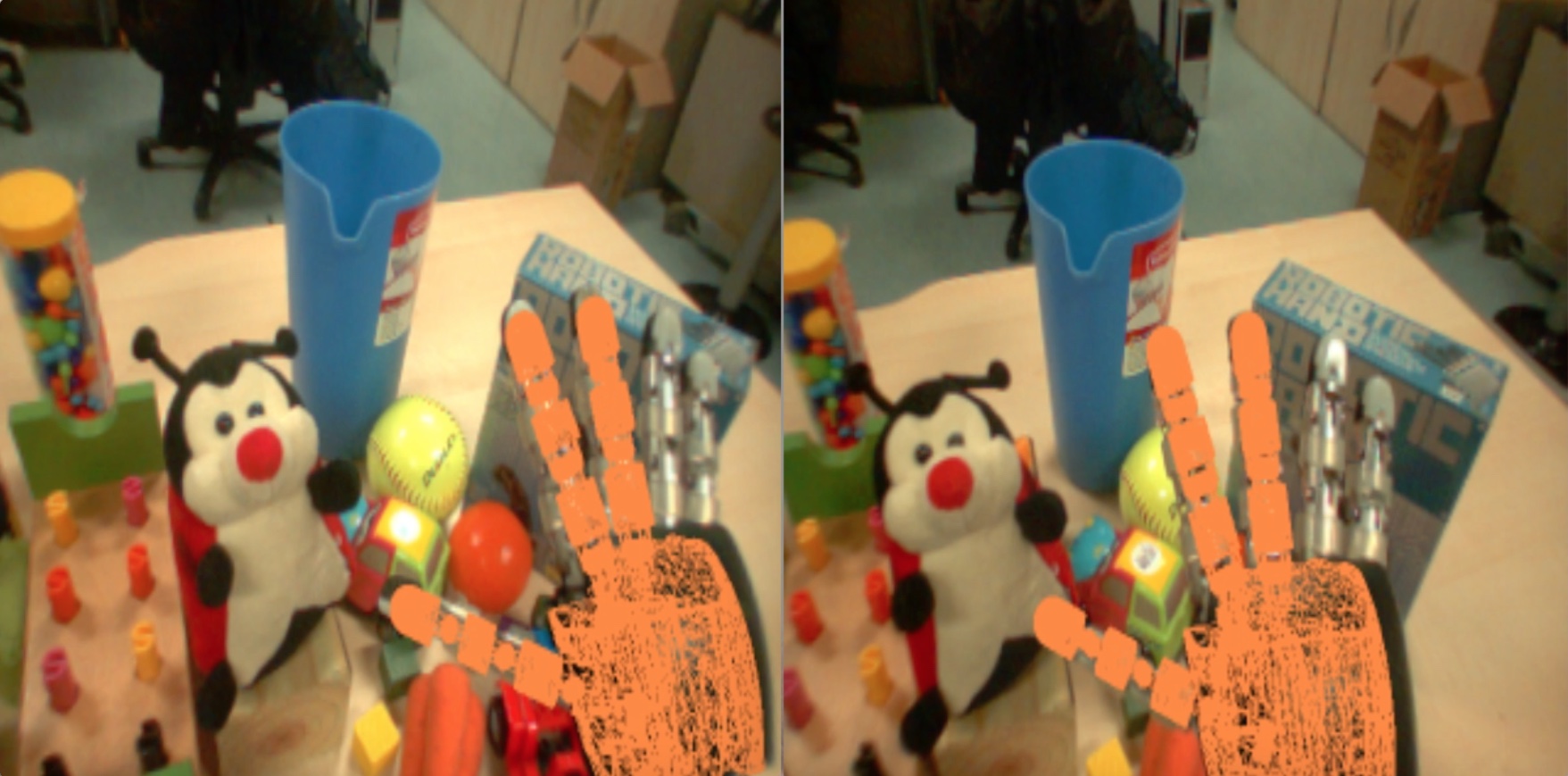}}}
        \caption{Filtered pose just before starting reaching motion.}
        \label{fig:reach2}\vspace{0.5em}
        \framebox{\parbox{0.97\linewidth}{\includegraphics[width=\linewidth]{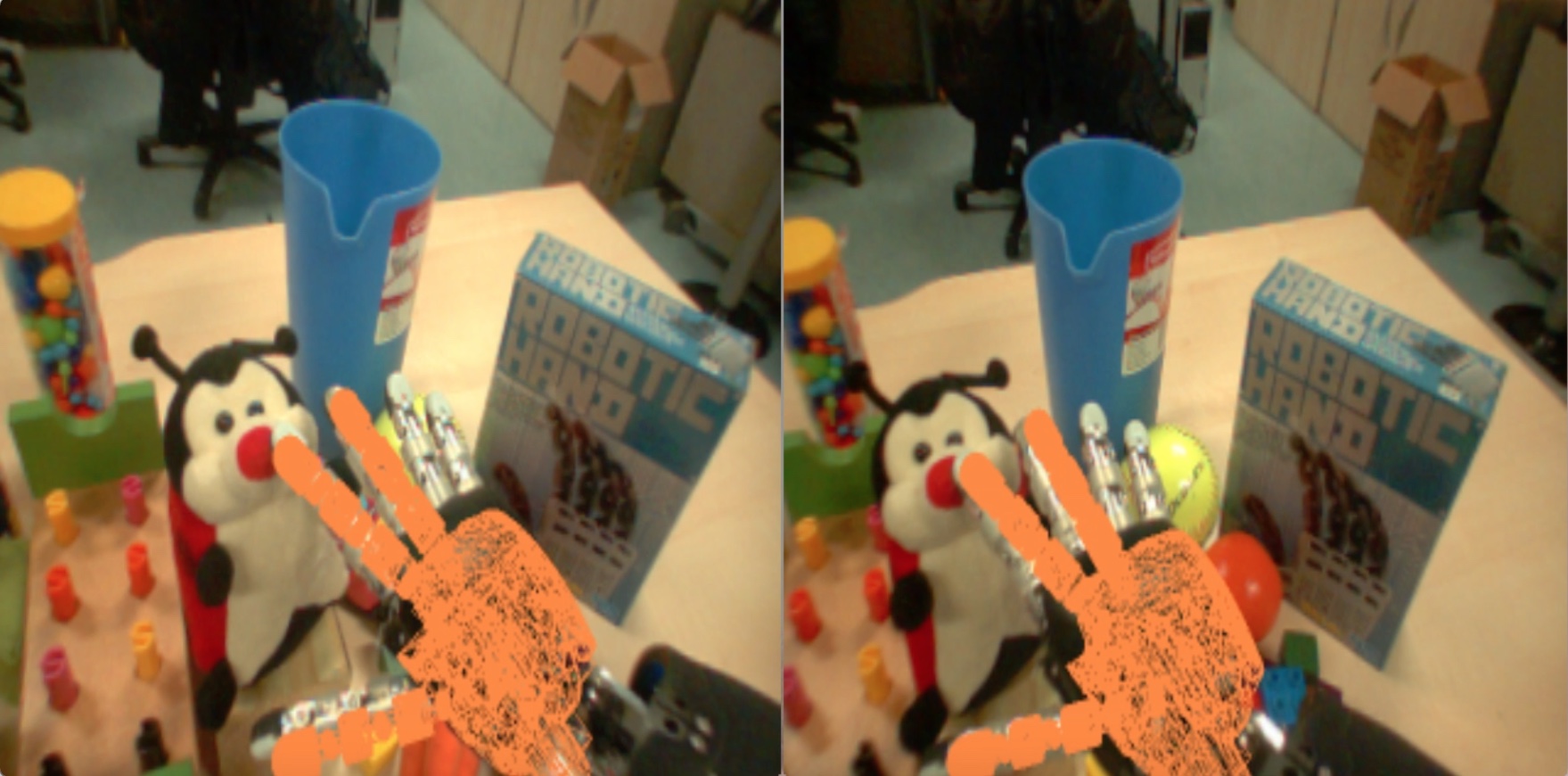}}}
        \caption{Filtered pose when reaching is completed.}
        \label{fig:reach3}\vspace{0.5em}
\end{figure}

Reaching was successful during the $10$ attempts: the filter turns out to be capable of tracking the hand pose while performing the reaching motion in a cluttered scenario.
Figs. \ref{fig:reach1}, \ref{fig:reach2} and \ref{fig:reach3} show, respectively, a snapshot of the initial discrepancy between the end-effector pose provided by the direct kinematics and the real one, the precision of the estimated pose before starting the reaching motion and the final position of the end-effector with the associated estimated pose.


\section{CONCLUSIONS AND FUTURE WORK}
\label{sec:con}
This paper tackles the estimation of the pose of a humanoid robot end-effector by means of a recursive Bayesian algorithm.
CAD models of the robot mechanical structure were used within a 3D rendering engine to virtually create images showing the manipulator as if they had been seen by the robot's cameras.
It was shown that 6D (position and orientation) estimation using dense Histogram of Oriented Gradient (HOG) descriptors on both camera and rendered images provide robust and reliable results.
The tracking algorithm has been integrated in a closed-loop control, to demonstrate that it allows to servo the hand with pixel accuracy using visual feedback.
This demonstrates that the algorithm can be effectively used in closed-loop to compensate for errors in the kinematic model of the robot and depth estimation.

The proposed approach paves the way to a number of possible future work comprising:
I$\left.\right)$ joint estimation of the pose of the end-effector and of the object to manipulate using, e.g., the random Random Finite Set (RFS) filtering approach \cite{mah2007,vo2010,hos2012,hos2013,mah2014,ris2016};
II$\left.\right)$ use data fusion techniques to exploit multiple features to improve both performance and robustness of the recursive Bayesian filtering;
III$\left.\right)$ conduct tests on different robot platform like, e.g., WALK-MAN \cite{tsa2016};
IV$\left.\right)$ distribute a Free and Open Source Software (FOSS) implementation of the algorithms to the community.


\section*{ACKNOWLEDGMENT}
This work was supported by the WALK-MAN project, funded under the European Community's 7th Framework Programme No. 611832 (Cognitive Systems and Robotics FP7-ICT-2013-10).


\addtolength{\textheight}{-2.7cm}   

\bibliographystyle{ieeetr}
\bibliography{IEEEabrv,iros_2017}

\end{document}